\documentclass{article}

\PassOptionsToPackage{numbers, compress}{natbib}
\usepackage[preprint]{neurips_2020}

\usepackage[utf8]{inputenc} 
\usepackage[T1]{fontenc}    
\usepackage{hyperref}       
\usepackage{url}            
\usepackage{booktabs}       
\usepackage{amsfonts}       
\usepackage{nicefrac}       
\usepackage{microtype}      
\usepackage[subtle,mathspacing=normal,mathdisplays=tight,tracking=normal]{savetrees}

\usepackage{amssymb,amsmath}
\usepackage{txfonts}
\usepackage{microtype}
\usepackage{xspace}
\usepackage{xcolor}
\xspaceaddexceptions{\%}
\usepackage{comment}
\usepackage{xcolor}
\usepackage{paralist}
\usepackage{graphicx}
\usepackage{subcaption}
\usepackage{rotating}
\usepackage{pdflscape}
\usepackage{algorithm}
\usepackage{algorithmicx}
\usepackage{algpseudocode}
\usepackage{multirow}
\DeclareMathOperator{\st}{s.t.}

\DeclareMathOperator*{\argmin}{arg\,min}

\newcommand{\keypoint}[1]{\textbf{#1}\quad}

\newcommand{\cut}[1]{}

\usepackage[capitalise]{cleveref}
\newcommand*\rot{\rotatebox{90}}

\usepackage{subfiles} 

\title{Don’t Wait, Just~Weight: Improving Unsupervised Representations by Learning Goal-Driven Instance Weights}

\author{%
  Linus Ericsson \\
  School of Informatics\\
  University of Edinburgh\\
  Edinburgh, UK \\
  \texttt{linus.ericsson@ed.ac.uk} \\
  \And
  Henry Gouk \\
  School of Informatics\\
  University of Edinburgh\\
  Edinburgh, UK \\
  \texttt{hgouk@inf.ed.ac.uk} \\
  \And
  Timothy M.~Hospedales \\
  School of Informatics\\
  University of Edinburgh\\
  Edinburgh, UK \\
  \texttt{t.hospedales@ed.ac.uk} \\
}

\begin{document}

\maketitle

\begin{abstract}
In the absence of large labelled datasets, self-supervised learning techniques can boost performance by learning useful representations from unlabelled data, which is often more readily available. However, there is often a domain shift between the unlabelled collection and the downstream target problem data. We show that by learning Bayesian instance weights for the unlabelled data, we can improve the downstream classification accuracy by prioritising the most useful instances. Additionally, we show that the training time can be reduced by discarding unnecessary datapoints. Our method, BetaDataWeighter is evaluated using the popular self-supervised rotation prediction task on STL-10 and Visual Decathlon. We compare to related instance weighting schemes, both hand-designed heuristics and meta-learning, as well as conventional self-supervised learning. BetaDataWeighter achieves both the highest average accuracy and rank across datasets, and on STL-10 it prunes up to 78\% of unlabelled images without significant loss in accuracy, corresponding to over 50\% reduction in training time.
\end{abstract}

\section{Introduction}
In the modern internet age, obtaining vast amounts of data is easier than ever and, as curating and annotating large datasets is expensive and labour-intensive, a promising direction is training deep models from more readily available unlabelled data. In recent years, self-supervised pre-training methods have been developed that are beginning to approach the performance of their supervised counterparts~\citep{Kolesnikov2019RevisitingLearning}, thus making the training of machine learning models far less labour-intensive. However, when training such models on uncurated data, it is by definition impossible to know if all the unlabelled data is useful or relevant to the downstream task. The target domain is often shifted or narrower compared to the source domain,  which could result in self-supervised learning damaging performance rather than helping it~\citep{Caron2019UnsupervisedData, Ji2019InvariantSegmentation, Peng2019InvestigatingLearning}. Compared to the rich literature of dealing with domain shift in supervised learning~\citep{Patel2015VisualAdvances}, the problem of domain shift in unsupervised visual representation learning has received little attention. Proposed solutions have focused on ad hoc approaches that have marginal benefits~\citep{Peng2019InvestigatingLearning} or require significant resources that are not available to the majority of practitioners~\citep{Caron2019UnsupervisedData}.

We approach this problem from the perspective of learning instance weights, a set of parameters determining the importance of each example during training. The best weights are the ones that minimise the expected loss on the downstream supervised target domain, when re-using the representation learned in the weighted source domain. Towards this end, we develop a meta-learning algorithm \citep{Thrun1998LearningLearn} termed \emph{BetaDataWeighter} that infers the optimal weights for self-supervised learning on unlabelled source data, in order to optimise performance on a relatively small labelled problem downstream in the target domain. As the weights are learned, the most helpful source data will be given priority, while other source data that would be unnecessary or even actively damaging in terms of target domain performance will be down-weighted.

Since this down-weighted data contributes little to learning, there is potential to accelerate training by pruning such data from the source set. However, realising this is challenging, as a datapoint that seems useless now might become useful at a later stage of training. Therefore, pruning apparently useless datapoints too early could be a false efficiency. In order to manage pruning-based acceleration with the potential future value of the data, we introduce a Bayesian approach, termed BetaDataWeighter, that meta-learns a beta distribution over the value of each datapoint. This enables us to prune instances more effectively and safely, based on both their their estimated value as well as the certainty of that estimation.

To evaluate our method on a variety of target task types, we introduce a new cross-domain self-supervised learning benchmark based on the data from the Visual Decathlon challenge~\citep{Rebuffi2017LearningAdapters}. Our results show that goal-driven self-supervised learning with BetaDataWeighter proves more effective than related heuristic and meta-learning alternatives.

Our main contributions are fourfold. 
Firstly, we study the novel problem of learning instance weights for self-supervised pre-training in support of a supervised target task in a different domain. Secondly, we introduce BetaDataWeighter, an effective meta-algorithm for goal-driven self-supervised instance weighting. 
This underpins a Bayesian dataset pruning mechanism that can prune up to 78\% of unlabelled images in STL-10~\citep{Coates2011AnLearning} without significant loss in accuracy, leading to over 50\% reduction in training time.
Lastly, we show that when learning CNN features using BetaDataWeighter to re-weight ImageNet, we can improve downstream classification performance on a variety of tasks from Visual Decathlon, with similar outcomes shown on STL-10. Our method outperforms recent related weighting schemes based on both hand-designed heuristics and meta-learning, as well as vanilla non-weighted self-supervision---achieving both the highest average accuracy and rank across datasets.
\section{Related Work}
\keypoint{Unsupervised Learning}
The most common approaches to learning without labels include clustering
\citep{Xu2015AAlgorithms, Caron2018DeepFeatures} and autoencoding \citep{Kingma2014Auto-encodingBayes}. In self-supervised learning, labels are generated automatically from the data itself \citep{Jing2019Self-supervisedSurvey} and the model is trained to predict them, via a `\emph{pretext task}'. Many recent computer vision works   have defined such tasks, like context prediction of image patches \citep{Doersch2015UnsupervisedPrediction, Noroozi2016UnsupervisedPuzzles}, colouring in grayscale images \citep{Larsson2016LearningColorization, Zhang2016ColorfulColorization} or predicting the image given  transformations of it to encourage invariant representations  \citep{Dosovitskiy2014DiscriminativeNetworks}. \citet{Kolesnikov2019RevisitingLearning} evaluate a range of these algorithms and find that the rotation-prediction pretext task \citep{Gidaris2018UnsupervisedRotations} is the most effective on natural image datasets like ImageNet \citep{Deng2009ImageNet:Database}. The model for this task, termed RotNet, is trained to predict which rotation has been applied to an image from the range $\{0^\circ, 90^\circ, 180^\circ, 270^\circ\}$. We build on RotNet by learning instance weights on the source data to improve both the performance when transferring learned features to a target domain, and the training time of self-supervised learning. Our goal is similar to that of \citep{Caron2019UnsupervisedData}, who also study learning from uncurated datasets, by hand-designing a robust unsupervised pre-training algorithm. In contrast, we meta-learn the pre-training strategy (paramaterised by instance weights) to optimise the downstream task, as well as the pruning strategy to accelerate self-supervised pre-training.

\keypoint{Instance-weighting and Curricula} Our strategy is based on instance re-weighting, which has been widely studied in the context of boosting \citep{Freund1997ABoosting}, hard negative mining \citep{Malisiewicz2011EnsembleBeyond}, focal loss \citep{Lin2017FocalDetection} and more recently in meta-learning  \citep{Ren2018LearningLearning, Shu2019Meta-Weight-Net:Weighting}. In particular, instance-weighing approaches have been successfully applied  to deal with noisy labels and class imbalance in supervised learning \citep{Nigam2020ImpactSurvey,Shu2019Meta-Weight-Net:Weighting,Ren2018LearningLearning}, by down-weighting mis-labelled images and over-represented categories respectively. In contrast, we study instance weighting for the purpose of self-supervised learning on uncurated data. In this case there is no label noise or class imbalance, but source images may have varying relevance to the downstream target problem. The challenges of learning on uncurated data have been highlighted for semi-supervised learning \citep{Peng2019InvestigatingLearning, Oliver2018RealisticAlgorithms}.
We focus on closing the loop between self-supervised pre-training and downstream supervised learning by meta-learning the optimal self-supervised instance weights. We ameliorate the computational overhead of such instance-wise meta-learning by introducing an effective data pruning strategy to eliminate unhelpful source data from consideration. This is in contrast to other approaches to dealing with data of varying relevance  \citep{Peng2019InvestigatingLearning,Shu2019Meta-Weight-Net:Weighting,Ji2019InvariantSegmentation} which all still pay the cost to train on the full data.

Instance pruning has been studied in the literature  \cite{Lapedriza2013AreValuable}, notably in the context of core-set construction \cite{Sener2018ActiveApproach,Bachem2017PracticalLearning,Munteanu2018Coresets-MethodsAlgorithms,Tsang2005CoreSets}. However, with the notable exception of \citep{Sener2018ActiveApproach}, few contemporary studies have attempted to prune the data used for deep networks online during model training. 


\keypoint{Meta Learning} 
Our weight learning strategy is an instantiation of meta-learning \citep{Thrun1998LearningLearn,Hospedales2020Meta-LearningSurvey}, which has seen a large number of works in recent years \citep{Finn2017Model-AgnosticNetworks, Andrychowicz2016LearningDescent}.
The meta-learning algorithms most related to our method are L2RW \citep{Ren2018LearningLearning} and MetaWeightNet \citep{Shu2019Meta-Weight-Net:Weighting}. We learn weights on the data in a transductive way, akin to L2RW, while MetaWeightNet learns an inductive model for weights. All these methods focus on within-domain learning, while we uniquely look at meta-learning instance weighing to facilitate domain-transfer between unlabelled source data and a downstream supervised learning task. Compared to L2RW \citep{Ren2018LearningLearning}, we do not normalise weights per batch (since for our unlabelled data, we cannot guarantee the proportion of good and bad data in each batch). More importantly, we achieve greater efficacy due to our Bayesian weight that enables continual weight learning without premature convergence, eliminating the need to reset weights after each batch.

\section{Background} \label{sec:background}
Our goal is to solve a machine learning problem in a target domain for which we have a small set of clean labelled data. The labelled set is too small for training a deep network from scratch, so we aim to exploit self-supervised pre-training on a large unlabelled  dataset. The auxiliary data is potentially uncurated, meaning that we do not know its composition and relevance to the target problem.

Let $\mathcal{D}_{target}=\{\mathcal{D}^{train}_{target}\cup\mathcal{D}^{val}_{target}\cup\mathcal{D}^{test}_{target}\}$ be the small labelled and curated dataset, and $\mathcal{D}_{source}$ be the large unlabelled dataset with unknown and varying  relevance to $\mathcal{D}_{target}$ including partial or no overlap in categories and low-level statistics. Our goal is to train an unsupervised feature extractor on $\mathcal{D}_{source}$ such that the accuracy of a classifier using this feature extractor and trained on $\mathcal{D}_{target}$ is maximised. 
To deal with the unknown and varying usefulness of the source data, we  adaptively re-weight source instances according to their contribution to target model performance. Furthermore, to reduce the cost of this process, we will discard the most unhelpful points from the source set online during training.  

Let $f_\theta$ be a neural network feature extractor and $\mathcal{L}_{ss}$ be a self-supervised loss on the unlabelled source data, $\mathcal{D}_{source} = (x_s^{(1)}, x_s^{(2)}, \dotsc, x_s^{(n)})$. The downstream target task consists of labelled data {$\mathcal{D}^{train}_{target} = ((x_t^{(1)}, y_t^{(1)}), (x_t^{(2)}, y_t^{(2)}), \dotsc, (x_t^{(m)}, y_t^{(m)}))$} with loss $\mathcal{L}_{meta}$, (which is typically a supervised loss but could potentially be unsupervised). We want to solve the following bi-level optimisation problem for instance weights $w = (w^{(1)}, w^{(2)}, \dotsc,w^{(n)})$:
\begin{equation} \label{eq:bilevel}
    \begin{gathered}
        \min_{w} \quad \frac{1}{m} \, \sum_j^m \, 
        \mathcal{L}_{meta}((x_t^{(j)}, y_t^{(j)}), \theta^*) \\
        \st \quad \theta^* = \argmin_\theta \quad \frac{1}{n}\sum^n_{i}w^{(i)} \mathcal{L}_{ss}(f_\theta(x_s^{(i)}))
    \end{gathered}
\end{equation}
In the outer optimisation level, the instance weights $w$ are updated so as to minimise the loss of the downstream task. In the inner level the feature extractor, $\theta$, is updated to minimise the weighted loss on the source task. Next we will introduce an algorithm for approximately solving this problem.

\section{Learning goal-driven instance weights}
We now introduce BetaDataWeighter (BDW), an online solution to the problem defined in Section~\ref{sec:background}. It maintains a distribution over the weights for each example, from which stochastic weights are generated at each update. These instance-wise parameters determine the importance of their corresponding datapoints throughout training. As they are only used during training, they do not affect inference run-time.

\subsection{BetaDataWeighter} \label{sec:bdw}
\keypoint{Instance-Weighted Feature Learning}
During training on the source data, the parameters $\theta$ of our feature extractor  $f_\theta$ are updated using  gradient descent on the self-supervised loss, $\mathcal{L}_{ss}$. For each datapoint, $x_s^{(i)}$, we associate a \emph{data-weight} parameter, $w^{(i)}$. This parameter is used to scale the loss on $x_s^{(i)}$ during training. A mini-batch update of size $k$, given the corresponding data-weights, is given by
\begin{equation}
    \label{eq:model_update}
    \theta^\prime \gets \theta - \alpha \, \nabla_\theta \, \frac{1}{k} \, \sum_{i = 0}^k \, w^{(i)} \mathcal{L}_{ss}(f_\theta(x_s^{(i)})),
\end{equation}
where $\alpha$ is the learning rate for the network parameters. By setting $w^{(i)}=1$ for all $i$, we obtain a standard stochastic gradient update. By tuning weights $w^{(i)}$, we can prioritise data that contribute more to downstream performance, and reduce the influence of data that is unhelpful or damaging.

\keypoint{Bayesian Instance-Weighting}
Instead of point estimating the weights, $w^{(i)}$, we can store the parameters defining a probability distribution over each weight. The weight of a datapoint can be viewed as the probability $p^{(i)}$ that an oracle data selector would select this datapoint for training. One can directly attempt to learn this value in a deterministic fashion (i.e. $w^{(i)} = p^{(i)}$). This is how many existing re-weighting methods operate. Alternatively, because $p^{(i)}$ can be interpreted as a Bernoulli parameter, we can model our belief about its value with a beta distribution. This distribution is itself parameterised by two scalars, $a^{(i)}$ and $b^{(i)}$. In our algorithm these will be the parameters that the outer loop optimises.

When sampling from a beta distribution, $w^{(i)} \sim Beta(a^{(i)}, b^{(i)})$, we obtain a data-weight in the range $[0, 1]$. This means that no weight can be negative and neither can any  point have a massive weight, which will help stabilise training. In order to define a beta distribution, the parameters $a^{(i)}$ and $b^{(i)}$ must be strictly positive. To learn these using unconstrained optimisation, we paramaterise them as  $\log(a^{(i)})$ and $\log(b^{(i)})$. Initially, we have no knowledge of whether a particular datapoint is relevant or not, so we initialise $p^{(i)}$ to a uniform prior by setting $a^{(i)}$ and $b^{(i)}$ to one. As training progresses, we will learn whether the datapoint is useful or not and thus update our distribution over its probability by updating the parameters $a^{(i)}$ and $b^{(i)}$. Like many meta-learning algorithms, we use our target to validate the model at each iteration of training. The target loss will provide  higher order gradients for updating $a^{(i)}$ and $b^{(i)}$. As solving the entire inner optimisation problem from \cref{eq:bilevel} is prohibitively expensive and can lead to gradient stability problems, we approximate it by performing a single inner step.

\begin{algorithm}[t]
{
    \caption{BetaDataWeighter}
    \label{alg:meta_beta_data_weighter}
\begin{algorithmic}[1]
    \State {\bfseries Input:} Source set $\mathcal{D}_{source}$, meta-set $\mathcal{D}^{train}_{target}$, learning rates $\alpha, \eta$, batch-size $k$, max epochs $T$
    \State {\bfseries Output:} Model parameters $\theta$
    \State For all $x_s^{(i)} \in \mathcal{D}_{source}$ initialise $a^{(i)} = 1, b^{(i)} = 1$
    \State Initialise model parameters $\theta$
    \For{epoch $t$ from $1$ to $T$}
        \For{sampled mini-batch $\{x_s^{(i)}\}_{i = 0}^k$ from $\mathcal{D}_{source, t}$}
            \State Sample $w^{(i)} \sim Beta(a^{(i)}, b^{(i)})$ by reparameterisation trick
            \State $\theta^\prime \gets \theta - \alpha \, \nabla_\theta \, \frac{1}{k} \, \sum_i^k \, w^{(i)} \mathcal{L}_{ss}(x_s^{(i)}; \theta)$ \Comment{Get new model params from update}
            \State $a \gets a - \eta \, \nabla_a \, \mathcal{L}_{meta}(\mathcal{D}^{train}_{target}, \theta^\prime)$ \Comment{Update $a$ using meta-gradient}
            \State $b \gets b - \eta \, \nabla_b \, \mathcal{L}_{meta}(\mathcal{D}^{train}_{target}, \theta^\prime)$ \Comment{Update $b$ using meta-gradient}
            \State $\theta \gets \theta^\prime$
        \EndFor
        \State $\mathcal{D}_{source, t + 1} = \{x_s^{(i)} \in \mathcal{D}_{source, t} \, | \, CDF(\lambda; a^{(i)}, b^{(i)}) > \rho\}$ \Comment{Prune datapoints}
    \EndFor
\end{algorithmic}
}
\end{algorithm}

\keypoint{Learning Algorithm} We describe our algorithm with batch-size one for illustration. At each iteration we sample an instance weight $w^{(i)}\sim\operatorname{Beta}(a^{(i)},b^{(i)})$, and then speculatively update model parameters $\theta$ given this weight
\begin{equation}
    \label{eq:single_weight_model_update}
    \theta^\prime \gets \theta - \alpha \, \nabla_\theta \, w^{(i)} \mathcal{L}_{ss}(f_\theta(x_s^{(i)})).
\end{equation}
After this speculative update, we evaluate the resulting model parameters $\theta^\prime$ on the $\mathcal{D}^{train}_{target}$ set to get a meta-loss that measures how well the new model generalises to unseen data. If labelled data is available, this meta-loss, $\mathcal{L}_{meta}$, can be a measure of classification performance, such as the cross entropy loss. However, it is also possible to use a self-supervised loss instead. This is discussed further in \cref{subsec:meta-loss}. As $w^{(i)}$ was obtained via a sampling process, we are unable to get gradients for the distribution parameters. But by reparameterising the beta distribution according to \citet{Figurnov2018ImplicitGradients} or by using the  tools available in recent deep learning libraries \citep{Paszke2019PyTorch:Library, tensorflow2015-whitepaper}, we can sample from it in a differentiable way. We can thus acquire gradients for $a^{(i)}$, and $b^{(i)}$ during our backward pass,
\begin{align}
    \label{eq:a_gradient}
    a^{(i)} &\gets a^{(i)} - \eta \, \nabla_{a^{(i)}} \, \mathcal{L}_{meta}(\mathcal{D}^{train}_{target}, \theta^\prime) \\
    \label{eq:b_gradient}
    b^{(i)} &\gets b^{(i)} - \eta \, \nabla_{b^{(i)}} \, \mathcal{L}_{meta}(\mathcal{D}^{train}_{target}, \theta^\prime).
\end{align}
where $\eta$ is learning rate for the instance weights, which is set independently of the model learning rate.

Each iteration of our algorithm consists of a model update followed by an update on the beta distribution parameters. Note that only the distributions corresponding to the datapoints within the current batch can be updated. Thus it takes a full epoch to update all of the distributions once. In this sense, the outer loop of \cref{eq:bilevel} is being solved with a block coordinate descent method, whereas the inner loop is solved with stochastic gradient descent. Pseudocode for the full BetaDataWeighter method is provided in \cref{alg:meta_beta_data_weighter}.
We give the details of a deterministic version, DataWeighter, in the appendix.

\subsection{Meta-loss}\label{subsec:meta-loss}
The loss, $\mathcal{L}_{meta}$, used to evaluate the feature extractor $\theta$ on $\mathcal{D}^{train}_{target}$ does not have to be the same loss used for training the feature extractor. When we have labels for this set, we prefer to use a loss that estimates the performance of a classifier. One could train a linear model $g_\phi$ and then measure performance based on predictions of $g_\phi(f_\theta(\cdot))$. However, this inner model fitting should be as efficient as possible since a new  model will be needed for each update.
As such, we define a meta-loss that constructs a nearest centroid classifier~\citep{Hastie2009ThePrediction} (NCC) and measures the performance of $\theta$ using cross entropy,
\begin{equation}
    \begin{gathered}
        \mathcal{L}_{meta}(\mathcal{D}, \theta) = \frac{1}{|\mathcal{D}|} \sum_{(x^{(i)}, y^{(i)}) \in \mathcal{D}} \mathcal{L}_{CE}(y^{(i)}, g_{\phi^*}(f_\theta(x^{(i)}))) \\
        \st \quad \phi^* = \argmin_{\phi} \mathcal{L}_{NCC}(\phi; \mathcal{D}, \theta),
    \end{gathered}
\end{equation}
where $\mathcal{L}_{CE}$ and $\mathcal{L}_{NCC}$ are the cross entropy loss and nearest centroid classifier loss, respectively. The simple nature of NCC permits a closed form solution to the $\argmin$ operation, as popularised by \citet{Snell2017PrototypicalLearning}. In initial experiments we found this to produce higher downstream classification accuracy compared to using a self-supervised meta-loss. At each iteration we sample a K-way-N-shot episode from the target set\cut{in order to train the nearest centroid classifier,} as in Prototypical Networks~\citep{Snell2017PrototypicalLearning} and use this to train the NCC and evaluate the meta-loss.

\subsection{Pruning Unhelpful Data}
Similar to the L2RW algorithm of \citet{Ren2018LearningLearning}, our method introduces computational overhead. We perform one extra forward pass and backward pass on the target set batch, plus an additional backward pass when updating the weight parameters. Although our use of NCC lowers computation overhead, we still find that our method is $3 \times$ to $6 \times$ slower than standard training.

The goal of learning instance weights is to improve  downstream model performance, but a useful side-effect is that it can also be used to reduce the training time by discarding irrelevant data. We thus introduce a pruning process between every epoch where unhelpful data is discarded and no longer revisited in future epochs.

\begin{table}[t]
\centering
\caption{Autoencoding MNIST, FashionMNIST (FMNIST) and KMNIST with mixed source data.  Lowest losses and training times are in \textbf{bold}, ignoring the Oracle VAE as it knows the domains.}
\vspace{0.2em}
\resizebox{0.66\columnwidth}{!}{
\label{tab:mnist}
\begin{tabular}{llcccccc}
\toprule
           &  & \multicolumn{2}{c}{MNIST}                                    & \multicolumn{2}{c}{FMNIST}                                   & \multicolumn{2}{c}{KMNIST}                                   \\ \midrule
           &  & \multicolumn{1}{l}{Test loss} & \multicolumn{1}{l}{Time (s)} & \multicolumn{1}{l}{Test loss} & \multicolumn{1}{l}{Time (s)} & \multicolumn{1}{l}{Test loss} & \multicolumn{1}{l}{Time (s)} \\ \midrule
Oracle VAE &  & 178.95                        & \phantom{0}1,387                        & 150.05                        & 2,256                        & 292.19                        & 2,161                         \\
VAE        &  & 209.69                        & \phantom{0}6,945                        & 171.46                        & 6,876                        & 306.43                        & 5,409                         \\
DW VAE (ours)  &  & 193.26                        & 15,856                       & \textbf{171.05}               & 9,943                         & \textbf{301.11}               & 5,137                         \\
BDW VAE (ours) &  & \textbf{192.64}               & \phantom{0}\textbf{6,005}               & 171.13                       & \textbf{1,424}                & 302.70                        & \textbf{4,081}                \\ \bottomrule
\end{tabular}}
\end{table}

Arising from our Bayesian approach, we are able to incorporate uncertainty about the data-weights in the pruning strategy. We wish to ensure that discarding data does not change the model objective significantly, as this could impact the stability of the training process---a phenomenon that leads to overfitting~\citep{Bousquet2002StabilityGeneralization}, and that we do not suffer from premature pruning (e.g., of low mean but high variance distributions). 
However, if a large proportion of an instance's weight density lies below a small value, then we are highly likely to sample a low data-weight so we can be confident about discarding it. We introduce two hyperparameters: the density threshold $\rho$, and the CDF threshold $\lambda$. A datapoint will be pruned if more than $\rho$ of the probability density lies below $\lambda$. Given source data for epoch $t$, $\mathcal{D}_{source, t}$, the data for training the BetaDataWeighter on epoch $t + 1$ is therefore
\begin{equation}
    \mathcal{D}_{source, t + 1} = \{x_s^{(i)} \in \mathcal{D}_{source, t} \, | \, CDF(\lambda; a^{(i)}, b^{(i)}) > \rho\},
\end{equation}
where $a^{(i)}, b^{(i)}$ are the beta distribution parameters of $x_s^{(i)}$ and $CDF(\cdot)$ is the cumulative density function.

\section{Experiments}
In this section we describe the series of experiments we conduct to evaluate our algorithm, the BetaDataWeighter, referred to as BDW. We also include results from running its deterministic version DW.

\subsection{Analysing instance weights} \label{sec:mnist}
We start by applying our method in a simple setup where we can track the weights of data which we know to be in-domain. A minimal variational auto-encoder (VAE) \citep{Kingma2014Auto-encodingBayes} is trained on three domains: MNIST \citep{LeCun2010MNISTDatabase}, FashionMNIST \citep{Xiao2017Fashion-MNIST:LearningAlgorithms} and KMNIST \citep{Clanuwat2018DeepLiterature}. Each dataset is chosen in turn to serve as the target domain, with its 10,000 test images serving as the target test set $\mathcal{D}^{test}_{target}$. We select another 10,000 images from its training set to serve as the target train set $\mathcal{D}^{train}_{target}$. The remaining training images are combined with the images from the training sets of the other two domains to form the unlabelled source set, $\mathcal{D}_{source}$. This creates a mixed source set of 170,000 images (i.e., 50,000 from MNIST, 60,000 from FashionMNIST, and 60,000 from KMNIST). No labels are used in this experiment as all losses are based on image reconstruction. Source set domain labels are also not used, and the goal is for our BDW VAE to select the most relevant source data for learning in order to encode each target problem.

The architecture consists of a fully-connected encoder that maps the input to 100 hidden units and then to a single hidden unit bottleneck. The decoder mirrors the structure of the encoder. The model is trained for 100 epochs with SGD using a learning rate of $1 \times 10^{-4}$ and a batch-size of 64.

\begin{figure}[t]
    \centering
    \begin{subfigure}[b]{0.48\textwidth}
        \includegraphics[width=\textwidth]{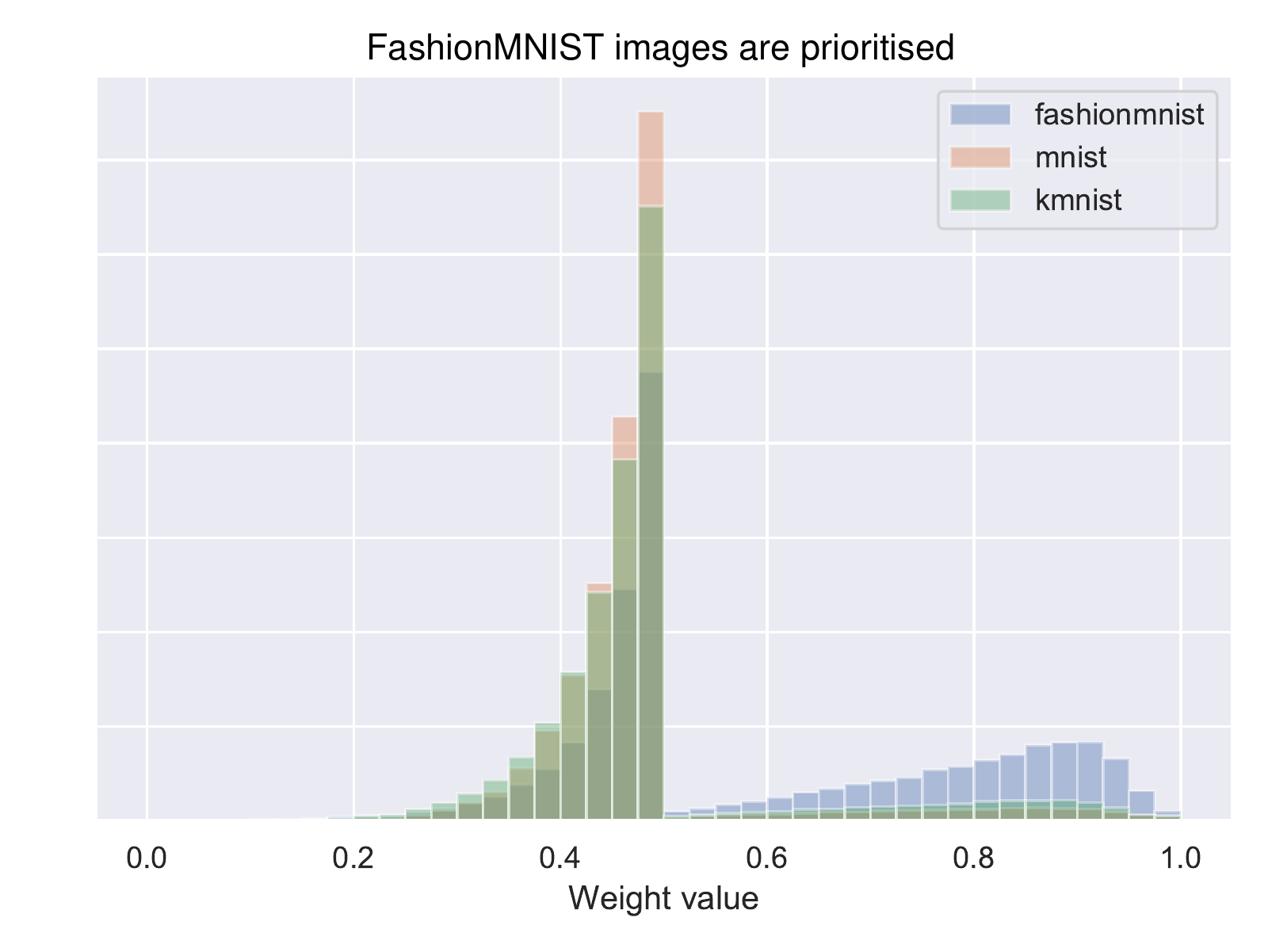}
        \caption{}
        \label{fig:mnist_weight_dist}
    \end{subfigure}
    ~ 
    \begin{subfigure}[b]{0.48\textwidth}
        \includegraphics[width=\textwidth]{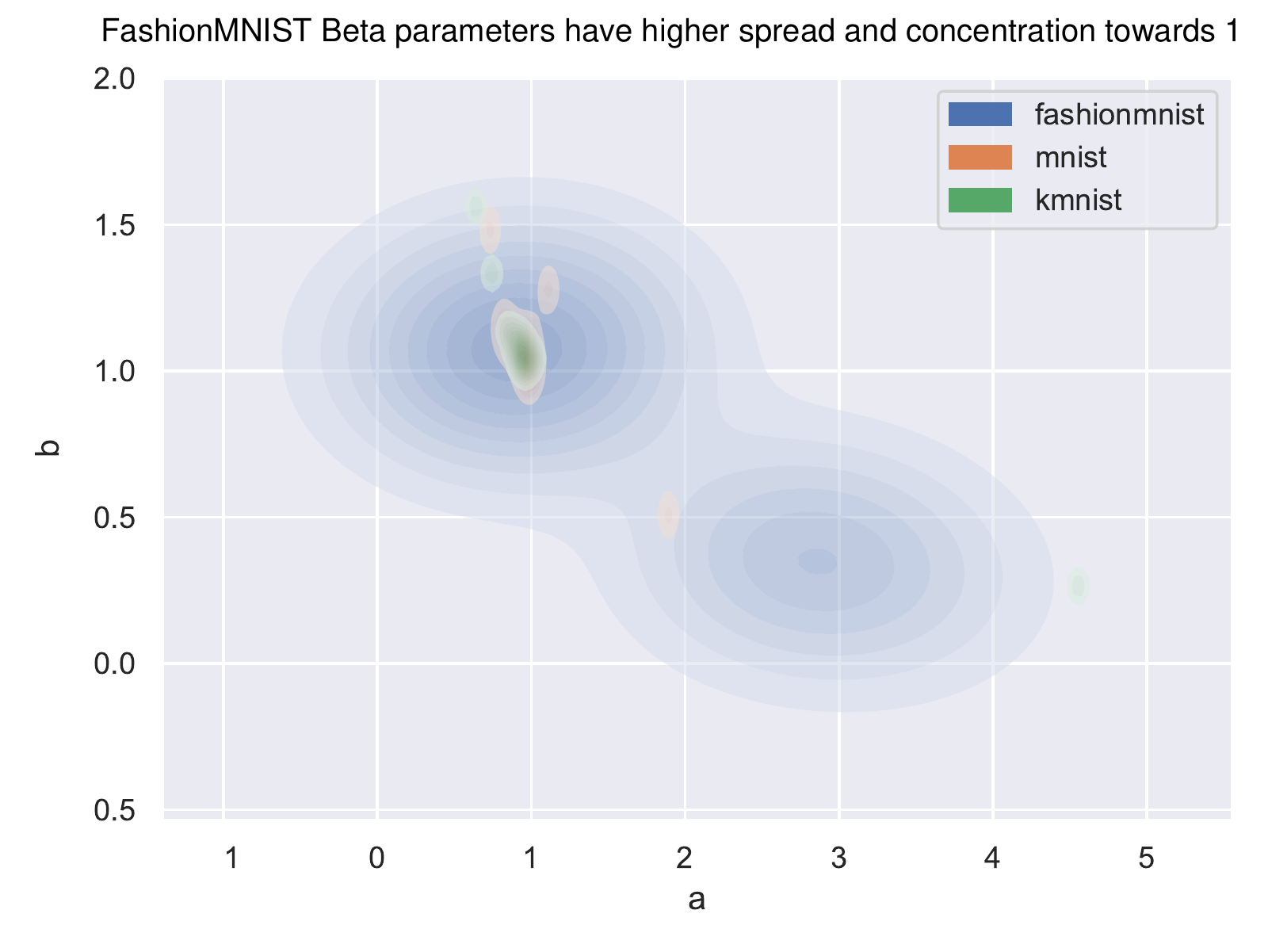}
        \caption{}
        \label{fig:mnist_beta_params}
    \end{subfigure}
    \vspace{-0.5em}
    \caption{(a) A histogram of the expected values of Beta distributions on source points on the FashionMNIST experiment. Data falling below the 0.5 line has been pruned. Blue: FashionMNIST, Orange: MNIST, Green: KMNIST. The majority of FashionMNIST images are still used in training and have high expected values while a majority of images from MNIST and KMNIST are pruned. (b) Density estimate of Beta parameters  $a$ and $b$. A high $b$ value means the density is concentrated towards 0, while a high $a$ means it's concentrated towards 1. Both figures were produced from the BetaDataWeighter run from \cref{tab:mnist} after epoch 25.}
\end{figure}

We evaluate two baselines: VAE, which trains on all three domains with all data-weights set to 1, and Oracle VAE, which only trains on the images from the target domain. Our BetaDataWeighter (BDW), and its deterministic ablation DataWeighter (DW), are trained on all of the data. \Cref{tab:mnist} shows that our methods improve the test loss compared to the baseline VAE which suffers from equally weighting source data of mixed relevance. Our methods perform similarly in terms of test loss, with both tending to improve on the VAE baseline. BDW manages to reduce the training time more significantly with better pruning.


\Cref{fig:mnist_weight_dist} plots the distribution of the learned data weights for the BetaDataWeighter with FashionMNIST target. It is clear that the images from FashionMNIST are prioritised over the other two domains. For the same run we also show the distribution of Beta parameters for all datapoints in \cref{fig:mnist_beta_params}. Here, a large number of FashionMNIST images tend to have higher $a$ and lower $b$ values. There is also a subset of FashionMNIST images that are have their density concentrated around lower weights, putting them in the same cluster of pruned datapoints as the MNIST and KMNIST data.

\subsection{Goal-Driven Self-Supervised Learning: Experiment Setup} \label{sec:vd}
We now investigate whether our method can improve the quality of visual features learned from a large dataset of unlabelled images, for transfer to a downstream classification task on a different domain. We use the Visual Decathlon (VD) \citep{Rebuffi2017LearningAdapters} collection of vision datasets. It consists of 10 very different image classification tasks; Aircraft, CIFAR100, DPed, DTD, Flowers, GTSRB, ImageNet, Omniglot, SVHN and UCF101. In our experiments on VD, the feature extractor is always trained on the 1.28 million images from the ImageNet domain as $\mathcal{D}_{source}$, and we use the remaining nine domains as downstream target tasks. As the test labels have not been released, we use the competition's validation splits as our test sets. For the nine target domains we create our own two sets from the competition train splits; $\mathcal{D}^{train}_{target}$ which is used to drive instance weight learning, and $\mathcal{D}^{val}_{target}$ which is used for model selection before evaluation on the test set. See details in the appendix for the size of the target splits. This benchmark is substantially larger than the previous experiment. As we saw in \cref{tab:mnist}, the BetaDataWeighter is often substantially faster than DataWeighter due to improved uncertainty-based pruning. 
We therefore focus on BetaDataWeighter on the VD experiments.

\begin{table}[t]
\centering
\caption{Test accuracies (\%) of ResNet34 on Visual Decathlon domains after meta-learning instance weights for self-supervised training on ImageNet. Similarly for STL-10. Best results are in \textbf{bold}.}
\label{tab:vd}
\resizebox{\textwidth}{!}{%
\begin{tabular}{llrrrrrrrrrr|r|r}
\hline
                                 &                                                   &  & \multicolumn{1}{c}{Airc.} & \multicolumn{1}{c}{C100} & \multicolumn{1}{c}{DPed} & \multicolumn{1}{c}{DTD} & \multicolumn{1}{c}{Flwr} & \multicolumn{1}{c}{GTSR} & \multicolumn{1}{c}{OGlt} & \multicolumn{1}{c}{SVHN} & \multicolumn{1}{c|}{UCF} & \multicolumn{1}{c|}{STL-10} & \multicolumn{1}{c}{Avg.} \\ \hline
\multirow{4}{*}{\rot{Log. reg.}} & RotNet                                            &  & 16.27                     & 11.36                    & 83.34                    & 11.43                   & 18.71                    & 71.17                    & 20.86                    & 50.67                    & 17.30                    & 45.65                       & 34.68                    \\
                                 & NN-Weighter \citep{Peng2019InvestigatingLearning} &  & 17.45                     & 12.66                    & 83.59                    & 11.54                   & 17.13                    & 72.10                    & \textbf{24.15}           & 54.72                    & 15.64                    & 45.68                       & 35.47                    \\
                                 & L2RW \citep{Ren2018LearningLearning}              &  & 16.70                     & 15.72                    & 85.36                    & 14.44                   & \textbf{30.27}           & \textbf{79.79}           & 18.71                    & 57.64                    & \textbf{22.89}           & 44.68                       & 38.62                    \\
                                 & BDW (ours)                                        &  & \textbf{21.13}            & \textbf{17.53}           & \textbf{86.38}           & \textbf{18.87}          & 24.48                    & 77.49                    & 23.49                    & \textbf{57.74}           & 22.49                    & \textbf{46.41}              & \textbf{39.60}           \\ \hline
\multirow{4}{*}{\rot{Finetuned}} & RotNet                                            &  & 19.74                     & \textbf{27.64}           & 95.00                    & 25.27                   & 16.67                    & 82.71                    & 71.47                    & 80.95                    & 29.25                    & 68.19                       & 51.69                    \\
                                 & NN-Weighter \citep{Peng2019InvestigatingLearning} &  & 19.62                     & 21.39                    & 95.82                    & 26.01                   & 37.65                    & 89.17                    & \textbf{72.78}           & 78.12                    & 28.43                    & 69.15                       & 53.81                    \\
                                 & L2RW \citep{Ren2018LearningLearning}              &  & 17.28                     & 15.45                    & 95.29                    & 18.67                   & 34.71                    & 89.65                    & 66.31                    & 79.05                    & 31.67                    & 63.13                       & 51.12                    \\
                                 & BDW (ours)                                        &  & \textbf{25.08}            & 24.90                    & \textbf{96.11}           & \textbf{28.72}          & \textbf{41.37}           & \textbf{93.96}           & 70.67                    & \textbf{82.21}           & \textbf{33.61}           & \textbf{71.12}              & \textbf{56.78}           \\ \hline
\end{tabular}%
}
\end{table}

\keypoint{Implementation details: Meta-Training}
We follow RotNet \citep{Gidaris2018UnsupervisedRotations} which is trained to predict which of four 2d rotations has been applied to an input image, from ($0^{\circ}, 90^{\circ}, 180^{\circ}, 270^{\circ}$).
Our network architecture is a ResNet34 \citep{He2016DeepRecognition}.
At each iteration, we sample a batch for the prototypical \cite{Snell2017PrototypicalLearning} style meta-loss. We use the common 20-way 5-shot design (20 classes with 5 examples of each) with some exceptions. See appendix for further details.

\keypoint{Implementation details: Evaluation} For final evaluation, we consider two cases: (i) Training a linear classifier on the fixed pre-trained features, and (ii) Fine-tuning the pre-trained representation on the target problem. For linear classifier evaluation: Features are extracted after the second residual block and a logistic regression model is fit using L-BFGS with the data from both the $D^{train}_{target}$ and $D^{val}_{target}$ sets.  For fine-tuning evaluation: We take the network which performed best on logistic regression and replace its classification head for the downstream task and finetune the entire network on the combined data from $\mathcal{D}^{train}_{target}$ and $\mathcal{D}^{val}_{target}$.

\keypoint{Baselines}
We compare our BetaDataWeighter to training the same rotation-prediction network on all data without any re-weighting. This baseline is referred to as \emph{RotNet}. The best feature extractor is selected by early stopping on the target validation accuracy computed with logistic regression.

We also implement the nearest neighbour weighting scheme in \citep{Peng2019InvestigatingLearning}. We refer to this baseline as the \emph{NN-Weighter}. Details of our implementation can be found in the appendix. The last baseline we compare to is the related meta-learning method \emph{L2RW} \citep{Ren2018LearningLearning}. L2RW is designed for within-domain supervised learning, but we adapt the algorithm for our task and use the same  prototypical style meta-loss loss on the $\mathcal{D}^{train}_{target}$ batch.


\keypoint{STL-10 experiment}
We also evaluate our method on STL-10 \citep{Coates2011AnLearning}, which contains 10 target supervised classes and has a larger amount of unlabelled out-of-domain images  for training. The 5000 labelled training images are split into our $\mathcal{D}^{train}_{target}$ and $\mathcal{D}^{val}_{target}$ sets, with 2500 images in each. For the meta-loss we use a 10-way 10-shot batch design. We train a ResNet18 using rotation-prediction on the unlabelled STL-10 images with a batch-size of 128 and extract features from the final pre-logit layer of the network. Random crops of $84 \times 84$ are extracted during training and centred crops of the same size for testing and finetuning. Other hyperparameters are identical to those in the ImageNet/VD benchmark. 

\subsection{Results}
\keypoint{Goal Driven Self-Supervised Learning}
The results of both experiments are shown in \cref{tab:vd}. From the results we can see that: (i) Weighting based methods usually improve on the vanilla self-supervised baseline of RotNet. Notably, (ii) BDW generally performs better than competitors. (iii) BDW is effective in both the scenario where a fixed feature is used to train a linear classifier, and where end-to-end fine-tuning is performed on the target problem. 

\begin{figure}[t]
    \centering
    \includegraphics[width=0.49\textwidth]{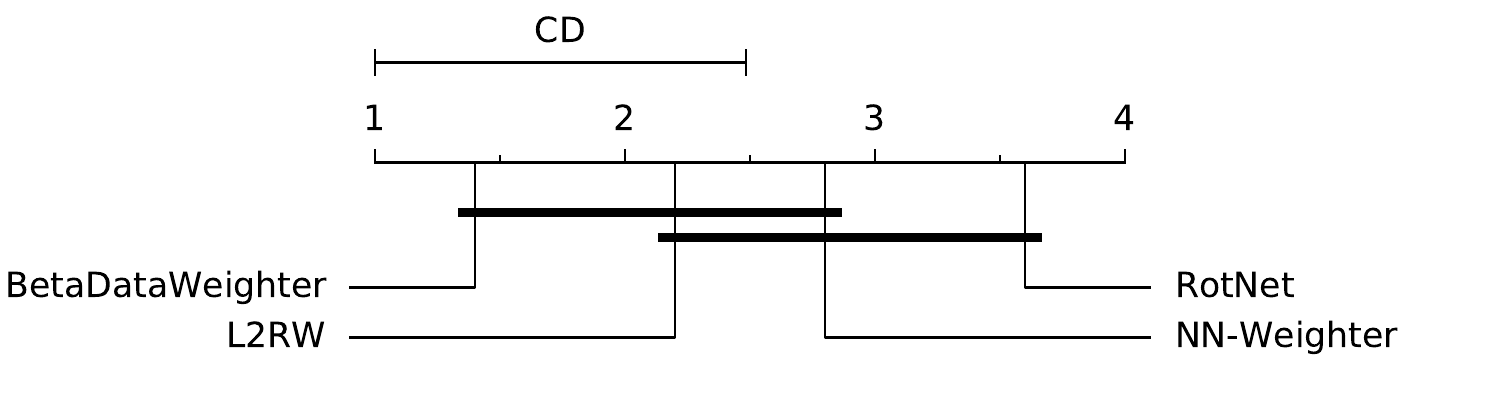}
    \includegraphics[width=0.49\textwidth]{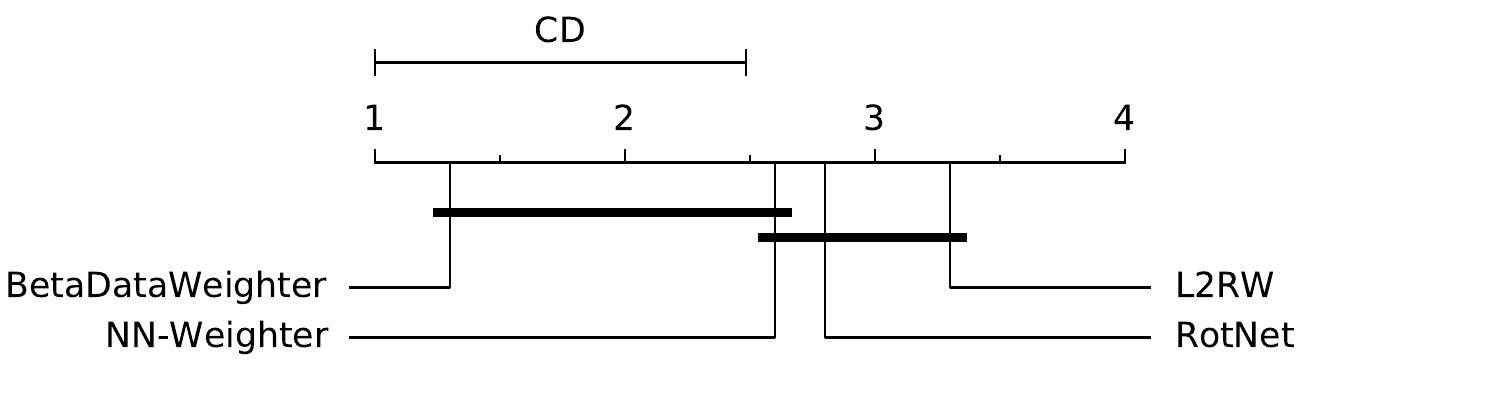}
    \caption{Critical difference diagrams from the Nemenyi test for the 10 datasets from VD and STL-10. Left: logistic regression on fixed features. Right: Finetuned. Each algorithm is represented by their average rank across all datasets. Bold lines connecting a group of algorithms indicate they are not significantly different as their average ranks differ by less than the CD value.}
    \label{fig:cd}
\end{figure}

To find out if the differences in performance of the models are statistically significant, we perform a Friedman test with $p = 0.05$ over all 10 domains in \cref{tab:vd}. The test passes so we reject the null hypothesis that all algorithms perform according to the same distribution. We follow this by a post-hoc Nemenyi test on the average ranks of the algorithms. The resulting CD diagrams in \cref{fig:cd}  confirm that the difference between RotNet and BetaDataWeighter is statistically significant both for the logistic regression results and finetuning scenarios.

\keypoint{Measuring the Effectiveness of Pruning}
To study the impact of pruning on accuracy and training time, we evaluate different pruning hyperparameters: outer learning rate ($\eta$),  density threshold ($\rho$) and the CDF threshold ($\lambda$), on the STL-10 (100,000 unlabelled and  5000 labelled images)  experiment. BetaDataWeighter is trained on the unlabelled images with labelled $\mathcal{D}^{train}_{target}$ and $\mathcal{D}^{val}_{target}$. From the results in \cref{tab:pruning} we can see that settings leading to both conservative and aggressive pruning achieve good accuracy. In particular, the finetuned test accuracies for all but one model are higher than the competing methods as seen in \cref{tab:vd}. This shows that a high pruning rate can reduce the training time while still boosting performance. However, unlike in our experiments from \cref{sec:mnist}, the training time never gets as low as the RotNet baseline. BDW does however improve upon the training time of the related meta-learning algorithm L2RW, showing the benefit of pruning.

\begin{table}[t]
\centering
\caption{BetaDataWeighter pruning on STL-10. A range of pruning levels achieve higher finetuned accuracy than the RotNet baseline. Training time can be significantly reduced compared to L2RW.}
\label{tab:pruning}
\resizebox{\textwidth}{!}{%
\begin{tabular}{llccccccccccccccc|cc}
\toprule
Learning rate            &  & \multicolumn{7}{c}{10}                                                                                                                                                                  &                      & \multicolumn{7}{c|}{100}                                                                                                                                                        & \multicolumn{1}{c}{} & \multicolumn{1}{c}{} \\ \cline{3-9} \cline{11-17}
Density threshold $\rho$ &  & \multicolumn{3}{c}{0.5}                                                              &                      & \multicolumn{3}{c}{0.9}                                                   &                      & \multicolumn{3}{c}{0.5}                                                     &                      & \multicolumn{3}{c|}{0.9}                                                   &                      &                      \\ \cline{3-5} \cline{7-9} \cline{11-13}  \cline{15-17} 
CDF threshold $\lambda$  &  & 0.1                                & \multicolumn{1}{c}{} & \multicolumn{1}{c}{0.25} &                      & \multicolumn{1}{c}{0.1} & \multicolumn{1}{c}{} & \multicolumn{1}{c}{0.25} &                      & 0.1                       & \multicolumn{1}{c}{} & \multicolumn{1}{c}{0.25} &                      & \multicolumn{1}{c}{0.1} & \multicolumn{1}{c}{} & \multicolumn{1}{c|}{0.25} & RotNet               & L2RW                 \\ \midrule
Finetuned                &  & \multicolumn{1}{r}{\textbf{71.53}} &                      & 71.12                    & \multicolumn{1}{r}{} & 70.74                   &                      & 70.09                    & \multicolumn{1}{r}{} & \multicolumn{1}{r}{69.40} &                      & 66.11                    & \multicolumn{1}{r}{} & 70.96                   &                      & 69.41                     & 68.19                & 63.13                \\
\% pruned                &  & \multicolumn{1}{r}{35.06}          &                      & 49.69                    & \multicolumn{1}{r}{} & 0.67                    &                      & 10.45                    & \multicolumn{1}{r}{} & \multicolumn{1}{r}{78.26} &                      & \textbf{83.56}           & \multicolumn{1}{r}{} & 47.49                   &                      & 68.38                     & 0.00                 & 0.00                 \\
Training time (s)        &  & \multicolumn{1}{r}{106,058}         & \multicolumn{1}{l}{} & 99,192                    & \multicolumn{1}{l}{} & 115,874                  & \multicolumn{1}{l}{} & 119,040                   & \multicolumn{1}{l}{} & \multicolumn{1}{r}{70,125} & \multicolumn{1}{l}{} & \textbf{53,931}           & \multicolumn{1}{l}{} & 101,552                  & \multicolumn{1}{l}{} & 86,143                     & 21,093                & 116,992               \\ \bottomrule
\end{tabular}%
}
\end{table}

\section{Conclusion}
Self-supervised representation pre-training on large unlabelled datasets is a popular strategy to boost supervised learning on smaller downstream tasks. We highlight the challenge posed by mixed-relevance source data, and introduce a meta-learning method to train self-supervised source instance-weights in support of the downstream goal. Our Bayesian approach of modelling distributions over weights leads to improved meta-learning efficacy and enables reliable source data pruning, which limits the computational cost of such meta-learning. Results across 10 target tasks show that our method outperforms alternatives. In future work we plan to apply this framework to also optimise the weighting of multiple self-supervised tasks.

\section*{Broader Impact}
In common with other unsupervised and self-supervised learning methods, our contribution promises to reduce the amount of manual effort required to annotate datasets for training machine learning models. These techniques can potentially benefit society by enabling small organisations with less data to more easily compete with large institutions holding huge datasets. Our method is particularly oriented at the most realistic case where auxiliary data is  unknown and of mixed relevance. While reducing manual labour is beneficial, in common with other unsupervised pre-training methods, we do trade-off this annotation effort for compute effort expended on the unlabelled set, which does entail some energy and environmental cost, and advantage those with greater access to compute resources. Our method ameliorates this issue to some extent with highly effective pruning techniques. These costs could be further reduced by future development in efficient meta-learning such as first order approximations and implicit differentiation. In future it may ultimately be possible to perform instance weighing with pruning more quickly than vanilla unweighted self-supervised learning.

\begin{ack}
This work was supported in part by the EPSRC Centre for Doctoral Training in Data Science, funded by the UK Engineering and Physical Sciences Research Council (grant EP/L016427/1) and the University of Edinburgh.
\end{ack}

\small
\bibliographystyle{custombib}
\bibliography{references,morerefs}

\newpage
\appendix
\section{Appendix} \label{sec:sup_mat}
\keypoint{Deterministic DataWeighter (DW)}
We present a deterministic version of our algorithm, where the data weights are point estimates, in contrast to the Bayesian approach in BDW described in \cref{sec:bdw}.

The data weights, $w$, are clipped to the range $[0, 1]$, similar to BDW. They are initialised to zero, like L2RW \citep{Ren2018LearningLearning}. During training these weights are directly optimised by using $\mathcal{D}^{train}_{target}$ to validate the model. The update for a single data weight $w^{(i)}$ becomes
\begin{align}
    \label{eq:ddw_w_gradient}
    w^{(i)} &\gets w^{(i)} - \eta \, \nabla_{w^{(i)}} \, \mathcal{L}_{meta}(\mathcal{D}^{train}_{target}, \theta^\prime)
\end{align}
where $\eta$ is learning rate for the data weights. The other properties of this algorithm are identical to BDW. Pseudocode for DW can be found in \cref{alg:deterministic_data_weighter}.

\begin{algorithm}[t]
{
    \caption{Deterministic DataWeighter (DW)}
    \label{alg:deterministic_data_weighter}
\begin{algorithmic}[1]
    \State {\bfseries Input:} Source set $\mathcal{D}_{source}$, meta-set $\mathcal{D}^{train}_{target}$, learning rates $\alpha, \eta$, batch-size $k$, max epochs $T$
    \State {\bfseries Output:} Model parameters $\theta$
    \State For all $x_s^{(i)} \in \mathcal{D}_{source}$ initialise $w^{(i)} = 0$
    \State Initialise model parameters $\theta$
    \For{epoch $t$ from $1$ to $T$}
        \For{sampled mini-batch $\{x_s^{(i)}\}_{i = 0}^k$ from $\mathcal{D}_{source, t}$}
            \State $\theta^\prime \gets \theta - \alpha \, \nabla_\theta \, \frac{1}{k} \, \sum_i^k \, w^{(i)} \mathcal{L}_{ss}(x_s^{(i)}; \theta)$ \Comment{Get new model params from update}
            \State $w \gets w - \eta \, \nabla_w \, \mathcal{L}_{meta}(\mathcal{D}^{train}_{target}, \theta^\prime)$ \Comment{Update $w$ using meta-gradient}
            \State $\theta \gets \theta^\prime$
        \EndFor
        \State $\mathcal{D}_{source, t + 1} = \{x_s^{(i)} \in \mathcal{D}_{source, t} \, | \, CDF(\lambda; a^{(i)}, b^{(i)}) > \rho\}$ \Comment{Prune datapoints}
    \EndFor
\end{algorithmic}
}
\end{algorithm}

\paragraph{Pruning in DW}
As DW does not have access to uncertainty of estimates, it prunes data solely based on the pruning parameter $\lambda$. If a data weight falls below this value, its associated datapoint is discarded. The source set at epoch $t + 1$ therefore consists of 
\begin{equation}
    \mathcal{D}_{source, t + 1} = \{x_s^{(i)} \in \mathcal{D}_{source, t} \, | \, w^{(i)} > \lambda\}.
\end{equation}

\begin{table}[t]
\centering
\caption{Combined size of the target train and validation splits.}
\resizebox{0.7\columnwidth}{!}{
\label{tab:splits}
\begin{tabular}{@{}lrrrrrrrrrr@{}}
\toprule
          & Airc. & C100 & DPed & DTD  & Flwr & GTSR & OGlt  & SVHN & UCF  & STL-10 \\ \midrule
\# images & 2000  & 2000 & 2000 & 1880 & 1020 & 2150 & 16230 & 2000 & 2020 & 5000   \\ \bottomrule
\end{tabular}}
\end{table}

\paragraph{Implementation Details: Size of Target Domain Splits}
For most domains we assign roughly 1000 images to each of these sets. However, in Flowers there are only 1020 images available so so we split them evenly across the two sets. In Omniglot there are a high number of classes (1623), so we assign 5 examples per class to each set. The combined sizes of these sets are given in \cref{tab:splits}.

\paragraph{Implementation Details: Meta-Training on VD domains}
We train each rotation-prediction model, as recommended by \citep{Kolesnikov2019RevisitingLearning}, for 35 epochs with SGD. In each batch, all four rotations of the image are included, effectively increasing the batch-size by a factor of four. It uses an initial learning rate of 0.1 which is decayed by 0.1 at epochs 15 and 25 and has a momentum parameter of 0.9. The batch-size is 256 (64 images $\times$ 4 rotations). All VD images have been resized to $72 \times 72$ and during training we take random crops of $64 \times 64$ pixels and reflect the image with a 50\% probability.
On some benchmarks we need to adapt the batch design as there are too few classes or too few examples of each class; on Daimler Pedestrian Classification we use 2-way 50-shot, on Omniglot 50-way 2-shot and on SVHN we use 10-way 10-shot classification.

\paragraph{Implementation Details: Meta-Testing on VD domains}
For linear classifier training, we use the hyperparameters suggested by \citep{Kolesnikov2019RevisitingLearning}, with 800 maximum iterations and the regularisation coefficient set to $\frac{100}{CM}$ where $C$ is the number of classes for this task and $M$ is the dimension of the extracted features.

For fine-tuning, the optimiser is Adam \citep{Kingma2015Adam:Optimization} with a batch-size of 64. We tune the initial learning rate, learning rate schedule and weight decay parameters using a held out set before re-tuning the network with the chosen hyperparameters. For both conditions we take a center crop of $64 \times 64$ and do not use any data augmentation.

\paragraph{Implementation Details: NN-Weighter}
This section describes our implementation of the nearest neighbour algorithm introduced by \citet{Peng2019InvestigatingLearning}. For every image $x^{(i)}$ in the training set, the method finds the Euclidean distance $d^{(i)}$ to the single nearest $\mathcal{D}^{train}_{target}$ image. It converts this distance into a weight, $w^{(i)} = \exp(-\beta d^{(i)})$, given hyperparameter $\beta$ which controls how quickly the weight approaches zero as the distance increases. As we use larger image spaces and datasets than the original authors we perform approximate nearest neighbour search using the HNSW method \citep{Malkov2016EfficientGraphs} within the NMSLIB Python package \citep{Boytsov2013EngineeringLibrary}. We tune $\beta$ on CIFAR100 and use the optimal value of $1 \times 10^{-5}$ across the whole range of domains.

\end{document}